\documentclass[journal]{IEEEtran}
\usepackage[nocompress]{cite}
\usepackage{url}
\usepackage{amsmath,amssymb,graphicx}
\usepackage{times}
\usepackage{epsfig}
\usepackage{makecell}
\usepackage{stackengine}

\ifCLASSINFOpdf
\else
\fi
\begin{document}
%
\title{Learning to Sort Image Sequences via Accumulated Temporal Differences}
%
%
%

\author{Gagan Kanojia and Shanmuganathan Raman

\thanks{Gagan Kanojia is with Electrical Engineering, Indian Institute of Technology Gandhinagar, Gandhinagar 382355, India (e-mail:gagan.kanojia@iitgn.ac.in).}
\thanks{Shanmuganathan Raman is with Electrical Engineering and Computer Science and Engineering, Indian Institute of Technology Gandhinagar, Gandhinagar 382355, India (e-mail:shanmuga@iitgn.ac.in).}
\thanks{The work of Gagan Kanojia was supported by the TCS Research Fellowship. The work of Shanmuganathan Raman was supported by an IMPRINT-2 grant}}

%
%

\markboth{Journal of \LaTeX\ Class Files,~Vol.~14, No.~8, August~2015}%
{Shell \MakeLowercase{\textit{et al.}}: Bare Demo of IEEEtran.cls for IEEE Journals}
%



\maketitle

\begin{abstract}
Consider a set of $n$ images of a scene with dynamic objects captured with a static or a handheld camera. Let the temporal order in which these images are captured be unknown. There can be $n!$ possibilities for the temporal order in which these images could have been captured. In this work, we tackle the problem of temporally sequencing the unordered set of images of a dynamic scene captured with a hand-held camera. We propose a convolutional block which captures the spatial information through 2D convolution kernel and captures the temporal information by utilizing the differences present among the feature maps extracted from the input images. We evaluate the performance of the proposed approach on the dataset extracted from a standard action recognition dataset, UCF101. We show that the proposed approach outperforms the state-of-the-art methods by a significant margin. We show that the network generalizes well by evaluating it on a dataset extracted from the DAVIS dataset, a dataset meant for video object segmentation, when the same network was trained with a dataset extracted from UCF101, a dataset meant for action recognition. All the codes and pretrained models will be publicly available at \_.
\end{abstract}


%
\IEEEpeerreviewmaketitle

\section{Introduction}
In today's world of digital photography, when a group of people attend an event like sports, they are likely to capture their moments of interest. These moments are generally short in duration and are dynamic in nature as they involves some moving objects or moving people present in the scene. Analysis of a dynamic scene using still images has been an active area of research in image processing, computer vision, and machine learning for a long time. However, the most common device for capturing such event is the mobile phone, which is a hand-held device. Even when the images are captured with a single handheld device, they are prone to misalignment due to reasons like handshakes. This makes the problem even more challenging, because apart from dealing with the object motion, the analysis also has to deal with the camera motion. This is due to the fact that the temporal information of the dynamic scene is an important tool for its analysis and visualization. In case the images are obtained from sources like internet, there may be no time stamp available. This makes the analysis of dynamic scenes extremely challenging. In \cite{sevilla2019only}, the authors showed that the temporal ordering plays an important role in recognizing several classes from the standard action recognition datasets\cite{kay2017kinetics,goyal2017something}.\\
In the past few years, 2D convolutional neural networks (CNNs) have been dominating several domains of computer vision like object recognition \cite{he2016deep}, single image depth estimation \cite{zhou2017unsupervised}, and semantic segmentation \cite{he2017mask}. There are several 2D CNN architectures which are being fueled with large still image datasets like ImageNet \cite{deng2009imagenet}. Apart from still images, it has also been shown that the 2D CNNs perform quite well when applied on videos \cite{xu2015discriminative,girdhar2017actionvlad,tran2017convnet}. They are applied on individual frames of the video to perform tasks such as action recognition. However, 2D CNNs lack in exploiting the 3D structure present in the input. To cope up with this issue, researchers moved on to the 2.5D approach which exploits the 3D structure while utilizing 2D convolution kernels \cite{alkadi20182,roth2014new}. In 2.5D approach, the network is provided with some higher level information about the input apart from the RGB channels present in the images. For example, in the case of action recognition, the higher level information could be optical flow and in the case of dynamic object detection, it could be semantic maps.\\

\noindent\textbf{Problem Statement.} Consider a set of $n$ images captured from single or multiple hand-held uncalibrated cameras whose order of capture, i.e., the temporal order, is unknown. In this work, we tackle a challenging problem of image sequencing in which we recover the unknown temporal order of the unordered set of images. Similar to \cite{lee2017unsupervised}, we formulate the problem of image sequencing as a classification problem and the classes are all the possible permutations of the temporal order of the given sequence length.  The objective is to map the given unordered image sequence to its corresponding permutation. Consider an input unordered image sequence $\mathcal{I} = \{I_2,I_4,I_1,I_5,I_3\}$ of length 5 whose correct order is $\{I_1,I_2,I_3,I_4,I_5\}$. The input sequence can have $n!$ possible permutations. In this work, we will consider the forward and backward permutations as a single class similar to \cite{lee2017unsupervised,kanojia2019deepimseq}. Hence, for a sequence of length 5, there are 5!/2 = 60 classes. The objective is to map $\mathcal{I}$ to its correct permutation.\\

\noindent\textbf{Contributions.} In this work, we propose a novel convolutional block for the task of image sequencing which extracts the spatial information using 2D convolution and the temporal information by exploiting the differences between the feature maps extracted from the input images. We do not provide any higher level information such as depth map and semantic information as an input. We only utilize the raw RGB images as the input. We use ResNet \cite{he2016deep} as the back-bone architecture for the proposed convolutional block. We show that the proposed approach outperforms the state-of-the-art methods by a significant margin. The proposed approach can be used as a pre-processing step in cases when the images of dynamic scene are obtained with no time stamp from sources like internet or a group of people \cite{kanojia2020simultaneous,kanojia2019patch,zarrabi2018crowdcam,dafni2017detecting}. In \cite{sevilla2019only}, the authors have already shown that the action recognition accuracy drops for several classes when the frames are randomly shuffled. In \cite{lee2018motion}, the authors have shown that even with 3 or 5 frames extracted from a video, a significant accuracy can be obtained for the task of action recognition. Also, in case of dynamic object detection and/or removal, the recent works have used around six images in the input set \cite{kanojia2020simultaneous,kanojia2019patch,zarrabi2018crowdcam,dafni2017detecting}. Hence, we limit our experiments up to six images which is also consistent with the recent works in image sequencing \cite{lee2017unsupervised,kanojia2019deepimseq}. \\
The major contributions of the work are as follows. 
\begin{itemize}
	\item  We propose a novel convolutional block which captures spatial information by performing a 2D convolution and temporal information by exploiting differences between the feature maps extracted from the unordered input images.
	\item We show that motion plays a key role in image sequencing through the motion heat maps computed using the outputs of the proposed block.
	\item We show that the network learns to shift its focus on dynamic objects without being trained with any such supervision by demonstrating the progression of motion heat maps along the depth of the network.
	\item We show that the network generalizes well by evaluating it on the dataset extracted from the DAVIS dataset, a dataset meant for video object segmentation, when the network is trained with the dataset extracted from UCF101, a dataset meant for action recognition.
	\item We outperform the state-of-the-art accuracy on the standard dataset used in previous works.
\end{itemize}
The rest of the paper is organized as follows. Section \ref{sec:related} discusses the previous works relevant to this work. Section \ref{sec:conv_block} describes the proposed convolutional block in detail. Section \ref{sec:experiments and discussions} discusses the network architecture and the experiments which show the effectiveness of the proposed method. It also discusses the ablation studies performed on the proposed convolutional block to justify the design choices. Section \ref{sec:conclusion} provides the conclusion.
\begin{figure}[t]
	\centering
	\includegraphics[width=\linewidth]{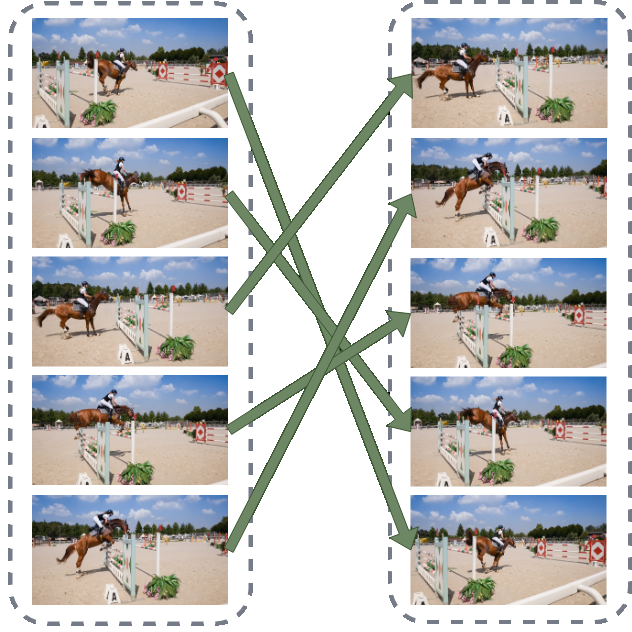}
	\caption{\textbf{Image sequencing.} The figure demonstrates the task of image sequencing.}
	\label{fig:im_seq_demo}
\end{figure}
\begin{figure*}[htbp]
	\centering
	\includegraphics[width=0.9\linewidth]{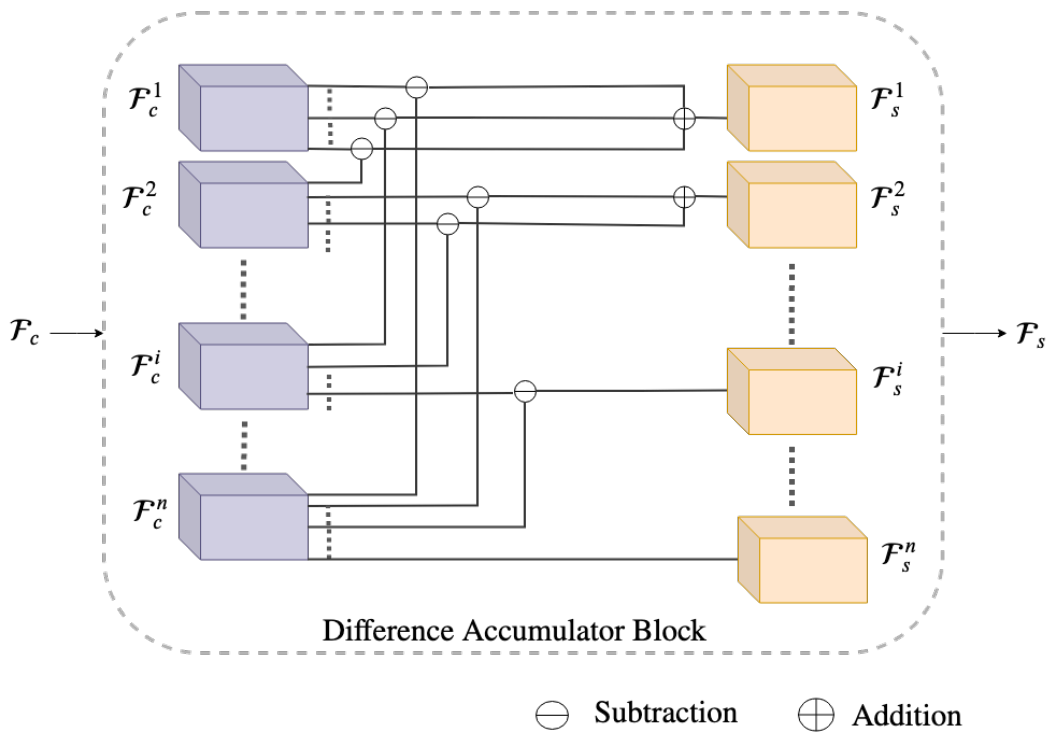}
	\caption{The figure shows an illustration of the difference accumulator block.}
	\label{fig:DAB}
\end{figure*}
\section{Related Work}
\label{sec:related}
In the past few years, 2D CNNs have enjoyed a huge amount of attention and have shown very promising results in several tasks of computer vision such as image classification, object detection, and image segmentation \cite{he2016deep,he2017mask}. They have been very successful in obtaining rich representations for still images. Many works extended 2D CNNs to operate on spatio-temporal data by extracting features of individual frames and then integrating the information along the temporal dimension \cite{xu2015discriminative,girdhar2017actionvlad}. In \cite{karpathy2014large}, the authors study different approaches to incorporate the temporal dimension along with the spatial dimensions through a ``slow fusion" model which extends the connectivity of the convolutional layers along the temporal dimension.\\
In many applications, the temporal structure of the input plays a very important role. In such cases, sequencing an unordered image sequence could help in better exploitation of temporal information \cite{baker2014learning,zhou2015temporal,huang2016visual,pickup2014seeing}. In \cite{zhou2015temporal}, the authors learn to predict the future actions of a person in an ego-centric video by performing two tasks related to the temporal reasoning among which one is the temporal ordering of the two given short video snippets. In \cite{huang2016visual}, the authors investigate whether the video is being played in the forward direction or the backward direction.\\
The problem of sequencing has been addressed in several scenarios like temporal ordering of the events in news \cite{mani2005temporally}, photo album creation from jumbled set of images \cite{sadeghi2015learning}, and estimating the 2-D rotation applied on the images to improve the feature representations \cite{noroozi2016unsupervised}. In \cite{misra2016shuffle}, the authors learn the video representations by learning to determine whether the input video is in correct temporal order or not. In \cite{agrawal2016sort}, instead of using only the images, the authors used image-caption pairs of an event of sequence length 5 and sorted them to make a story.\\
The problem of sequencing images of a dynamic scene captured by a hand-held camera was first addressed by Basha \emph{et al.} \cite{basha2012photo,moses2013space}. In their work, in the case of multiple cameras, they assume that they know the cluster of images belonging to the same camera and they know the temporal order of the images captured from the same camera. Also, they assume that at least two images are captured from almost the same location. In \cite{kanojia2014shot}, the authors deal with these assumptions by proposing a methodology when given a set of images captured from multiple cameras, they cluster the images captured with the same camera. After clustering, they sort the images temporally in their order of capture. However, these methods are not learning-based approaches. Also, they are not evaluated on large datasets.\\
The recent works by Lee \emph{et al.} \cite{lee2017unsupervised} and Kanojia \emph{et al.} \cite{kanojia2019deepimseq} are the most relevant to us. Lee \emph{et al.} \cite{lee2017unsupervised} propose a learning-based approach to sort an unordered image sequence. They formulate the task of sequencing as a multi-class classification problem in which the classes are all the possible permutations of the temporal order of the input unordered image set. However, they do not feed the images directly to the network. Given the images, they extract image regions which exhibits large motion and then feed these regions to the network along with some pre-processing. Unlike Lee \emph{et al.} \cite{lee2017unsupervised}, Kanojia \emph{et al.} \cite{kanojia2019deepimseq} directly feed the images into their proposed LSTM-based network. They formulated the task of image sequencing as a sequence-to-sequence mapping task. Their proposed network maps the input images to its position in the ordered sequence.\\
The proposed approach uses only 2D convolution kernels. Since 2D convolution kernels lack in capturing the temporal information, we adopt a 2.5D approach. In 2.5D approach, the RGB channels of the image are appended by some higher level information which captures the temporal structure of the input \cite{alkadi20182,roth2014new}. In \cite{roth2014new}, the authors fuse the input image with its orthogonal views. In \cite{alkadi20182}, the authors extend the dimension of the magnetic resonance volume along the RGB channels to exploit the 3D features. In the proposed approach, we extract the differences among the feature maps along the temporal direction and append them along the channels. Then, we perform the 2D convolution to extract the 3D structure present in the input unordered image set.\\
Temporal difference has been explored in the area of action recognition \cite{lee2018motion}. Temporal differences can provide the rough location of the non-rigid bodies performing the action in the videos \cite{park2013exploring,wang2009evaluation}. In \cite{lee2018motion}, the authors proposed a motion filter in which they compute the differences only among the feature maps of the adjacent frames. However, in our case, we do not have the information regarding the adjacency of the input frames. Also, in \cite{lee2018motion}, authors perform 1D convolution on the feature differences and then add them to the previous map while we adopt a 2.5D approach.
\section{Proposed Convolutional Block}
\label{sec:conv_block}
The proposed convolutional block has three parts: a 2D convolution kernel, a difference accumulator block (DAB), and a 2.5D convolutional block. Let the input feature map to the convolutional block be $\mathcal{F} \in \mathbb{R}^{c\times n \times h \times w}$. Here, $c$ is the number of channels, $n$ corresponds to the number of images in the input set, and $h$ and $w$ are the height and width of the feature map $\mathcal{F}$, respectively.
\subsection{2D Convolution}
It has been shown in the early works \cite{zeiler2014visualizing, krizhevsky2012imagenet}, that in the initial layers, the 2D filters learn to capture the information like edges and corners. In the later layers, they try to capture object-level information of the scene. The idea behind performing 2D convolution is to first obtain rich spatial representations individually. In the proposed framework, first we obtain $\mathcal{F}_c$ as shown in Eq. \ref{eq:1d_conv}.
\begin{equation}
\mathcal{F}_c = \mathcal{F}\star h_c
\label{eq:1d_conv}
\end{equation}
Here, $\star$ stands for convolution and $h_c$ is the filter for 2D convolution of kernel size $1\times k \times k$ and $c$ channels. The reason behind representing the kernel size of the filter $h_c$ with three dimension is to indicate that we are dealing with multiple images. This is mentioned for the sake of clarity that we are not performing convolution along the temporal direction. It can be seen that the kernel size along the temporal direction is $1$. We apply $h_c$ on the feature maps corresponding to each image, which are provided in the unordered fashion, to obtain their spatial feature maps. Then, we pass $\mathcal{F}_c$ through the proposed difference accumulator block (DAB) to obtain the temporal structure of the feature maps.
\subsection{Difference Accumulator Block}
\label{sec:DAB}
The core idea behind Difference Accumulator Block (DAB) is to capture the 3D structure present in the input data. Since we want to find the temporal order, we exploit the changes occurring among the images. The changes can be due to the object motion or the camera motion. In the proposed DAB, we rely on the differences among the feature maps extracted from the images, i.e., the change in the spatial information at different time instances, to extract the necessary temporal information present in the feature maps. These differences can provide rough locations of the non-rigid bodies performing action in the videos which could help the network in performing image sequencing \cite{park2013exploring,wang2009evaluation}. In a general sense, DAB tries to capture the notion of how different is the volume of feature maps at the current temporal location in comparison to the volumes at the other temporal locations.\\
Let $\mathcal{F}_c \in \mathbb{R}^{c\times n \times h \times w}$ be given as the input to DAB. Here, $c$ is the number of channels, $n$ corresponds to the number of images in the input set, and $h$ and $w$ are the height and width of the feature map $\mathcal{F}_c$, respectively. We pass $\mathcal{F}_c$ through DAB. We accumulate the differences of the feature map corresponding to each image of the unordered sequence with the features maps of the remaining images, i.e., the differences between the volumes along the second dimension of $\mathcal{F}_c$. Let $\{\mathcal{F}_c^1,\mathcal{F}_c^2, \mathcal{F}_c^3,\ldots,\mathcal{F}_c^n\} \in \mathbb{R}^{c\times 1 \times h \times w}$ be the volumes along the temporal depth of $\mathcal{F}_c$. The output of DAB $\mathcal{F}_s \in \mathbb{R}^{c\times n \times h \times w} $ is obtained as shown in Eq. \ref{eq:DAB}. Here, $\mathcal{F}_s$ is the concatenation of $\{\mathcal{F}_s^1,\mathcal{F}_s^2, \mathcal{F}_s^3,\ldots,\mathcal{F}_s^n \}\in \mathbb{R}^{c\times 1 \times h \times w}$ along the temporal depth.
\begin{equation}
\mathcal{F}_s^i = \sum\limits_{k=i+1}^{n}(\mathcal{F}_c^{i}-\mathcal{F}_c^{k}), \ \ \forall i = 2, \ldots, n-1
\label{eq:DAB}
\end{equation}
Here, $i$ is a location along the temporal dimension. For $i=n$, $\mathcal{F}_s^n=\mathcal{F}_c^n$. It can be observed that the range of $k$ starts from $i+1$. This is performed to avoid symmetric computations. Fig. \ref{fig:DAB} shows an illustration of the proposed DAB.\\
\subsection{2.5D Convolutional Block}
To capture the temporal information along with the spatial information, we adopt 2.5D approach. In 2.5D approach, the channels containing the spatial information are appended by some higher level information which captures the temporal structure of the input \cite{alkadi20182,roth2014new}. In our case, $\mathcal{F}_s$ obtained from DAB contains essential information regarding the temporal structure of the input image set. The feature maps $\mathcal{F}_c$ obtained by applying the 2D convolution kernel contain only the spatial information. To exploit both the spatial and the temporal structure of the input, we concatenate $\mathcal{F}_c$ and $\mathcal{F}_s$ along the channels to obtain $\mathcal{F}_{sc}\in \mathbb{R}^{2c\times n \times h \times w}$. Then, we pass $\mathcal{F}_{sc}$ through a 2D convolution kernel to obtain the final output of the block.
\subsection{Forward/Backward Propagation}
The forward and backward propagation through the proposed convolutional block is quite straightforward. The first component of the proposed convolutional block is a standard 2-D convolution filter through which gradients can be passed using standard backpropagation. The second component is DAB. In DAB, we extract the feature maps from the input feature maps by tensor slicing and then, perform subtraction and addition to obtain the output. These operations can be done in a differentiable manner using standard deep learning libraries. Finally, the third component is again a convolution kernel through which the gradients can be passed using standard backpropagation.
\section{Experiments and Discussions}
\label{sec:experiments and discussions}
\subsection{Datasets}
\subsubsection{UCF101}
UCF101 is a standard action recognition dataset which contains real world action videos \cite{soomro2012ucf101}. It contains 13320 videos out of which 9537 videos are used for training and 3783 videos are used for testing purpose. It is a diverse dataset which covers 101 action categories. The videos have large camera variations, cluttered background, and illumination conditions. It has been used as a benchmark dataset in several works such as \cite{tran2015learning,tran2017convnet,diba2018spatio}. Lee \emph{et al.} \cite{lee2017unsupervised} extract image sequences of length 3 and 4 from the training set of the split-1 of UCF-101. To extract the image sequences, they estimate optical flow in the videos and based on the magnitude of the optical flow, they extract the image sequences. Kanojia \emph{et al.} \cite{kanojia2019deepimseq} extend their dataset by including the image sequences of lengths 5 and 6. They obtain sequences of length 5 by extracting a frame from the left of the sequences of length 4. Similarly, they obtain sequences of length 6 by extracting the frames from the left and the right side of the sequences of length 4. While extracting the frames, the authors made sure that the temporal spacing between the frames is consistent with the original sequence of length 4. We randomly split the datasets corresponding to each sequence length into training and testing as 70\% and 30\% of the data, respectively. The videos of UCF101 contains 101 action categories which are grouped in 25 groups. Each group contains 4-7 videos. The video belonging to same group can contain common features. Hence, while splitting, we made sure that the image sequences belonging to the same group fall into the same category, i.e., training and test set.
\subsubsection{DAVIS}
DAVIS dataset is a benchmark dataset in the area of video object segmentation \cite{perazzi2016benchmark}. It contains fifty video sequences with several challenging scenarios like occlusions, appearance variation, and motion blur. The videos are captured with a moving camera. They contain single and multiple dynamic objects. We extract the dataset of evenly spaced image sequences of lengths 4 and 6 from the videos. We use this dataset to evaluate the generalization capability of the proposed approach for the task of image sequencing. We do not train the network on the sequences extracted from this dataset. We treat the whole dataset as the test set. We use the network trained on the dataset extracted from UCF101 to estimate the temporal order of the sequences extracted from the DAVIS dataset when provided in an unordered fashion.
\begin{figure}
	\centering
	\stackunder{\includegraphics[width=0.45\linewidth]{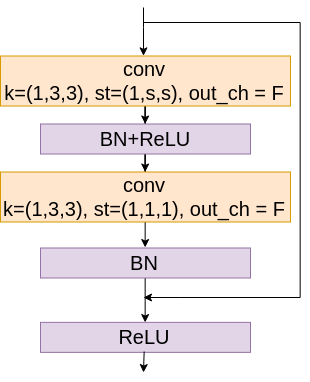}}{(a) ResNet (Basic)}
	\stackunder{\includegraphics[width=0.45\linewidth]{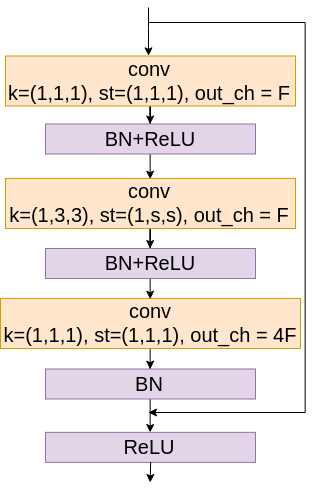}}{(b) ResNet (Bottleneck)}
	\stackunder{\includegraphics[width=0.48\linewidth]{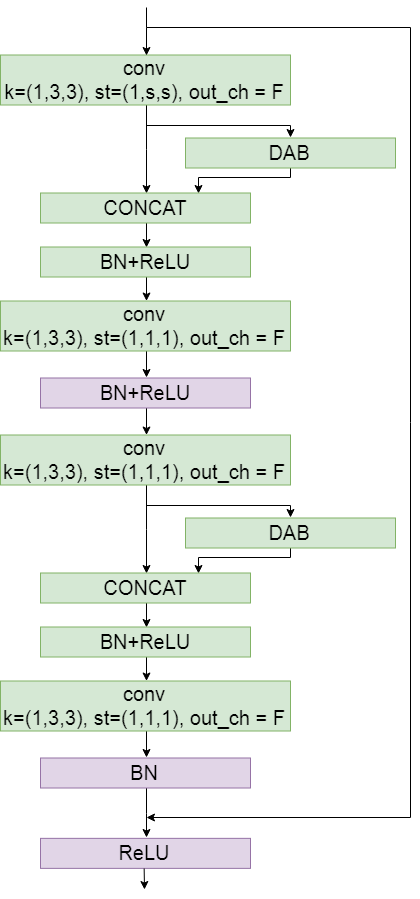}}{(c) Ours (Basic)}
	\stackunder{\includegraphics[width=0.48\linewidth]{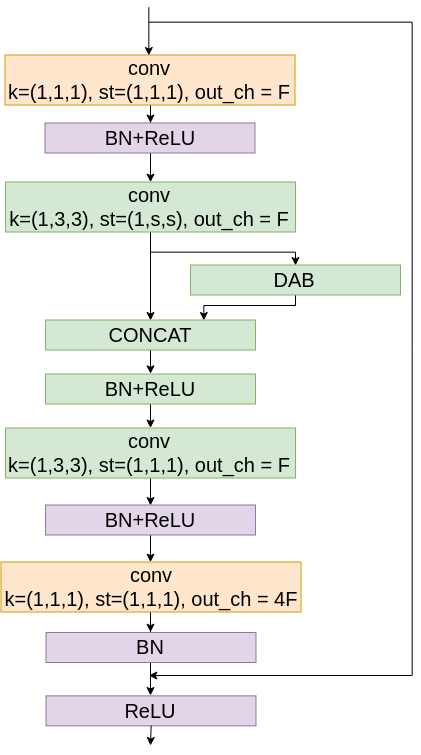}}{(d) Ours (Bottleneck)}
	\caption{(a) and (b) show the basic and bottleneck blocks used in ResNet architecture \cite{he2016deep}. (c) and (d) show the residual blocks in which 2D convolution kernel is replaced by the proposed convolutional block (in green). Here, $k$, $st$, $out\_ch$, $conv$, $BN$, and $ReLU$ stands for kernel size, stride, output channels, convolution, batch normalization and rectified linear unit, respectively.}
	\label{fig:blocks}
\end{figure}
\subsection{Network Architecture}
\label{sec:network_arch}
We use residual networks (ResNet) as the back-bone architecture to show the effectiveness of the proposed convolutional block \cite{he2016deep}. Fig. \ref{fig:blocks} (a) and (b) show the basic and bottleneck block used in ResNet architecture \cite{he2016deep}. In each of the residual blocks, we replace the 2D convolution kernel by the proposed convolutional block (in green) as shown in the Fig. \ref{fig:blocks} (c) and (d). We just replace the 2D convolution kernels by the proposed convolutional block while keeping the overall structure of the network intact. We perform the experiments with 18 layers and 50 layers version of the ResNet architecture. Similar to \cite{kanojia2019deepimseq}, we train separate networks for each sequence length. The input to the network is an unordered set of images of a dynamic scene captured with a hand-held camera. Similar to \cite{lee2017unsupervised} and \cite{kanojia2019deepimseq}, the forward and the backward permutations are considered as a single class. Hence, the classification layer of the network has $n!/2$ classes. Each class corresponds to a permutation. The objective of the network is to map the input unordered image sequence to its corresponding permutation.
\begin{figure*}[htbp]
	\centering
	\includegraphics[width=\linewidth]{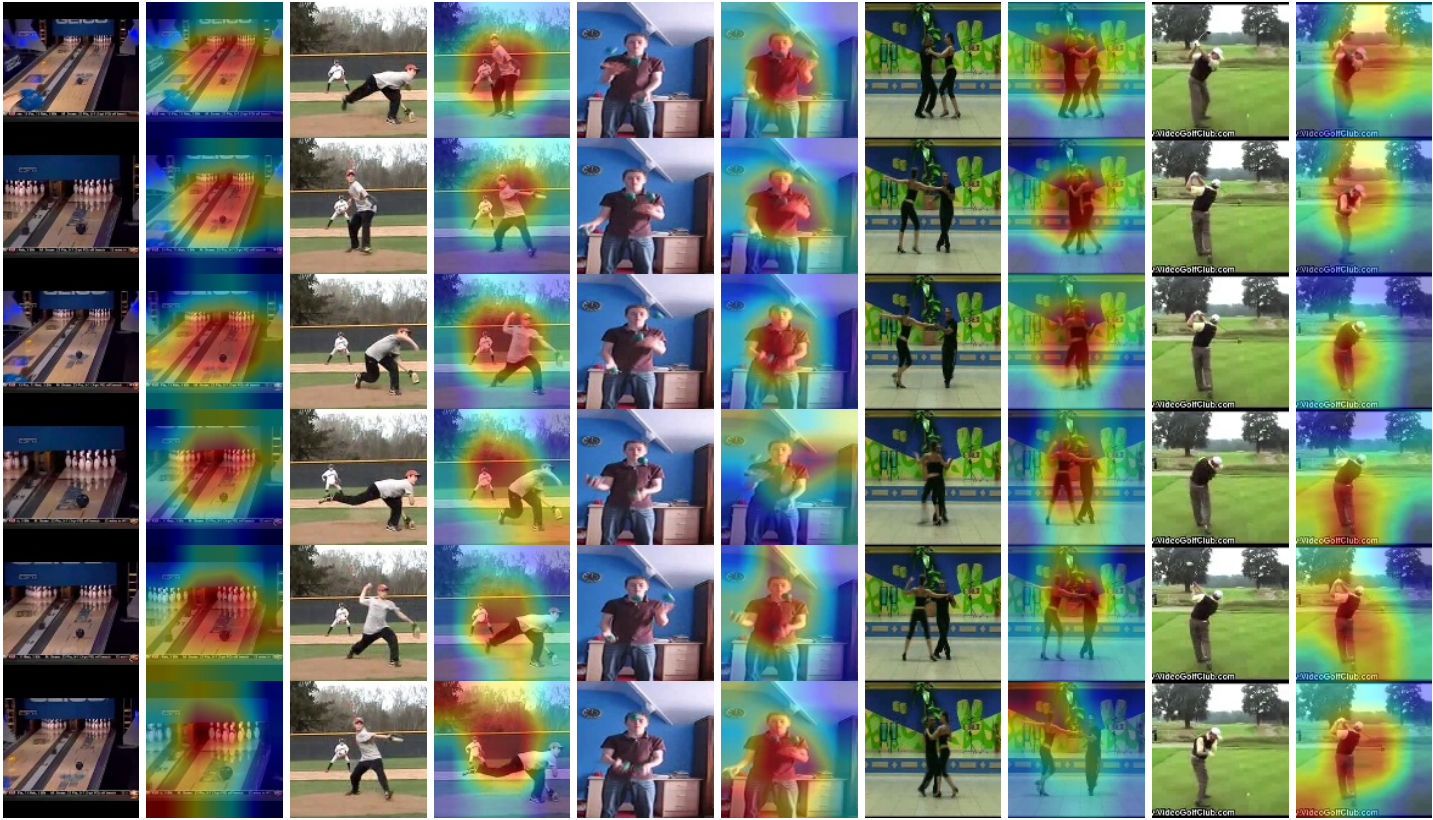}
	\caption{\textbf{Motion Heap Maps.} The figure shows the five test sets (first out of two columns in each set) of unordered image sequences extracted from UCF-101 dataset which has been correctly classified by the proposed network trained on the training set extracted from UCF-101. In each set, the first column shows an unordered image set which is given as the input to the proposed network and the second column shows the order of images provided by the proposed network as the output along with the motion heat maps computed from the output of the last DAB of the network.}
	\label{fig:ucf101}
\end{figure*}
\subsection{Training}
We train the networks with Stochastic Gradient Descent (SGD) for the weight update with a momentum of 0.9, weight decay of 0.001, and an initial learning rate of 0.1. We reduce the learning rate by 0.1 when the validation loss saturates. We use a batch of 16 clips for all the networks. The data augmentation used in training the networks is the same as that used in \cite{kanojia2019deepimseq}. Similar to \cite{kanojia2019deepimseq}, we perform a random cropping on the input image sets. The clips are spatially resized such that the shorter edge gets scaled to 136 pixels and then, we randomly crop a region of size $112\times 112$. The size of each data sample is $3\times n \times 112 \times 112$. Here, $3$ is the number of channels, $n$ is the number of images in the input sequence, and $112$ is the spatial size of the clips. We normalize the frames by subtracting the mean values and dividing them by the variance values of the ImageNet \cite{deng2009imagenet}. The training set for sequences of length 3, 4, 5, and 6 contains around 87.7K, 87.7K, 85K, and 83K sets of image sequences, respectively. We train the networks by feeding all the permutations of the image sequences in a random order. For example, to train the network for the sequence length 6, we feed the network with $83K \times 6!/2 \approx 29.9M$ unordered image sequences. Similarly, we test the networks with all the permutations of the images sequences in the test sets. We use categorical cross-entropy as the loss function.
\subsection{Comparisons with the State-of-the-art Methods}
Table \ref{tab:state_of_the_art} compares the test classification accuracy obtained on the datasets of unordered image sequences of different sequence lengths extracted from UCF-101 by the proposed approach with the state-of-the-art methods proposed by Lee \emph{et al.} \cite{lee2017unsupervised} and Kanojia \emph{et al.} \cite{kanojia2019deepimseq}. All the networks are trained from scratch. Table \ref{tab:state_of_the_art} shows the accuracy obtained using the proposed approach with ResNet (50 layers) as the backbone architecture. It can be observed that there is a significant improvement over the previous methods in terms of classification accuracy. The proposed approach outperforms the state-of-the-art method by Kanojia \emph{et al.} \cite{kanojia2019deepimseq} by a significant margin. It can be observed that the margin grows as we move from the sequence length of 3 to 6. This shows that the proposed approach is better in handling longer sequences in comparison to  Kanojia \emph{et al.} \cite{kanojia2019deepimseq} and Lee \emph{et al.} \cite{lee2017unsupervised}. Fig. \ref{fig:ucf101} shows the results on four test sets of unordered image sequences extracted from UCF-101 dataset obtained by the proposed network trained on the training set extracted from UCF-101.
\begin{table}[t]
	\centering 
	\caption{\textbf{Comparison with the state-of-the-art.} The table compares the test classification accuracy (in percentage) obtained on the datasets of unordered image sequences of different sequence lengths extracted from UCF-101 by the proposed approach with ResNet (50 layers) as backbone architecture with the state-of-the-art methods proposed by Lee \emph{et al.} \cite{lee2017unsupervised} and Kanojia \emph{et al.} \cite{kanojia2019deepimseq}.}
	\begin{tabular}{|c|c|c|c|c|} \hline
		\makecell{Sequence\\ Length} & Lee \emph{et al.}\cite{lee2017unsupervised} & Kanojia \emph{et al.}\cite{kanojia2019deepimseq} & Ours \\\Xhline{2\arrayrulewidth}
		3 & 63 & 67.18 & \textbf{83.04} \\
		4 & 41 & 60.33 & \textbf{80.10} \\
		5 & - & 54.78 & \textbf{81.85} \\
		6 & - & 51.30 & \textbf{78.29} \\
		\Xhline{2\arrayrulewidth}
	\end{tabular}
	\label{tab:state_of_the_art}
\end{table}
\subsection{Ablation study}
\subsubsection{Without DAB}
In this study, we verified the importance of DAB. The output of DAB is $\mathcal{F}_s$ which contains the temporal structure of the input obtained by accumulating the difference between the features extracted from the input image set. To check its importance, during training we set $\mathcal{F}_s=0$, i.e., we fill zeros at all positions in $\mathcal{F}_s$, in all the layers of the network. We keep everything else exactly the same. We experimented with the image sequences extracted from UCF-101 of length 3 and 4 with ResNet (18 layers) as the back-bone. We observed that the network does not learn anything and gives the output accuracy equivalent to the random probability which is $1/(n!/2)$ for a sequence of length $n$. This shows that extracting temporal structure is very crucial for the task of image sequencing. Also, this study confirms that DAB plays a significant role for the task.
\subsubsection{Effect of Network Depth}
In this study, we observe the effect of the depth of the back-bone network on the sequences of lengths 3 and 4. We used ResNet with 18 layers and 50 layers as the back-bone architecture for the proposed convolutional block. We replaced the 2D convolution kernel in the residual blocks with the proposed convolutional block. We train them with the datasets of image sequences extracted from UCF-101 of lengths 3 and 4. Table \ref{tab:depth_effect} shows the comparison of the test classification accuracy obtained on the datasets of unordered image sequences extracted from UCF-101 of lengths 3 and 4 when trained with the networks of different depth.  It can be observed in Table \ref{tab:depth_effect} that the deeper network (ResNet-50) performs better than the shallower network (ResNet-18).
\begin{table}[t]
	\centering 
	\caption{\textbf{Effect of network depth.} The table shows the comparison of test classification accuracy (in percentage) obtained on the datasets of unordered image sequences extracted from UCF-101 of lengths 3 and 4 when trained with the networks of different depths.}
	\begin{tabular}{|c|c|c|c|c|c|} \hline
		Sequence Length & Back-bone Network & Accuracy  \\\hline
		3 &  ResNet (18 layers) & 80.94  \\\hline
		3 &  ResNet (50 layers) & \textbf{83.04}  \\\hline
		4 &  ResNet (18 layers) &  77.79 \\\hline
		4 &  ResNet (50 layers) & \textbf{80.10} \\\hline
	\end{tabular}
	\label{tab:depth_effect}
\end{table}
\subsubsection{Effect of the Sign of Differences}
In this study, we observe the effect of the sign of the differences computed in DAB. To observe its effect, we modified Eq.~\ref{eq:DAB} of the block as shown in Eq.~\ref{eq:DAB_abs}.
\begin{equation}
\mathcal{F}_s^i = \sum\limits_{k=i+1}^{n}|(\mathcal{F}_c^{i}-\mathcal{F}_c^{k})|, \ \ \forall i = 2, \ldots, n-1
\label{eq:DAB_abs}
\end{equation}
Here, $|.|$ stands for the absolute value of the input. $\mathcal{F}_s^i$ is defined in Section~\ref{sec:DAB}. Instead of accumulating differences with their sign, we only accumulate their magnitude. We train the network with ResNet (18 layers) as the back-bone on the dataset of unordered image sequences of sequence length 4 extracted from UCF-101 with the modified DAB, i.e., with only the magnitude of the differences. Table~\ref{tab:effect_of_sign} shows the comparison of the test classification accuracy obtained on the dataset of unordered image sequences extracted from UCF-101 of sequence length 4 by the proposed approach when we train the network with DAB comprising Eq.~\ref{eq:DAB} with that of when it is trained with DAB comprising Eq.~\ref{eq:DAB_abs}. It can be observed that the network performs significantly better with both the sign and the magnitude.
\begin{table}[t]
	\centering 
	\caption{\textbf{Effect of Sign.} The table shows the comparison of the test classification accuracy (in percentage) obtained on the dataset of unordered image sequences extracted from UCF-101 by the proposed approach when the network is trained with the difference accumulator block comprising Eq.~\ref{eq:DAB} with when it is trained with the difference accumulator block comprising Eq.~\ref{eq:DAB_abs}. ResNet (18 layers) is used as back-bone architecture used for this experiment.}
	\begin{tabular}{|c|c|c|c|c|c|c|} \hline
		& \makecell{Sequence Length}  & Accuracy  \\\hline
		Magnitude & 4 & 60.82   \\\hline
		Sign+Magnitude& 4 & \textbf{77.7}  \\\hline
	\end{tabular}
	\label{tab:effect_of_sign}
\end{table}
\begin{table}[h!]
	\centering 
	\caption{\textbf{Effect of varying the number of images for accumulating differences.} The table shows the comparison of test classification accuracy (in percentage) obtained on the dataset of unordered image sequences of length 5 extracted from UCF-101 when trained with different values of m (Eq. \ref{eq:DAB}).}
	\begin{tabular}{|c|c|c|c|c|c|} \hline
		m & Back-bone Network & Accuracy  \\\hline
		0 &  ResNet (50 layers) & 1.667  \\\hline
		1 &  ResNet (50 layers) & 1.667  \\\hline
		2 &  ResNet (50 layers) & 77.16 \\\hline
		n &  ResNet (50 layers) & \textbf{81.85} \\\hline
	\end{tabular}
	\label{tab:m_effect}
\end{table}
\begin{table}[t]
	\caption{\textbf{Generalizability.} The table shows the classification accuracy (in percentage) obtained on the dataset  of unordered image sequences of length 4 and 6 extracted from DAVIS dataset using the proposed approach (ResNet50 as the backbone) when the network is trained on the dataset extracted from the UCF-101. The first column shows the spacing (in terms of frames) between the images of the extracted image sequences in the original video. For each sequence length, i.e. 4 and 6, the first column shows the accuracy obtained on sets of unordered image sequences when they are extracted with the corresponding temporal spacing and the second column shows the number of permutations of image sequences extracted from the DAVIS dataset used for obtaining the classification accuracy. \#Perms stands for the number of all the permutations of image sequences.}
	\centering
	\begin{tabular}{|c|c|c|c|c|c|c|} \hline
		& \multicolumn{4}{c|}{Sequence Length} \\\hline
		& \multicolumn{2}{c|}{4} & \multicolumn{2}{c|}{6}   \\\hline
		step & Accuracy  & \#Perms  & Accuracy  & \#Perms  \\\hline
		1 & 92.99 & $\approx$ 0.142M & 90.7 & $\approx$ 4.146M \\\hline
		2 & 93.43 & $\approx$ 0.136M & 89.7 & $\approx$ 3.821M  \\\hline
		3 & 91.8 & $\approx$ 0.129M & 84.5 & $\approx$ 3.498M \\\hline
		4 & 87.61 & $\approx$ 0.123M & 77.7 & $\approx$ 3.173M  \\\hline
		5 & 84.04 & $\approx$ 0.116M & 72.3 & $\approx$ 2.849M   \\\hline
		6 & 80.90 & $\approx$ 0.110M & 68.2 & $\approx$ 2.529M  \\\hline
		7 & 77.35 & $\approx$ 0.103M & 64.6 & $\approx$ 2.209M    \\\hline
		8 & 74.74 & $\approx$ 0.097M & 62.2 & $\approx$ 1.898M  \\\hline
	\end{tabular}
	\label{tab:davis}
\end{table}
\begin{figure}[htbp]
	\centering

	\includegraphics[width=\linewidth]{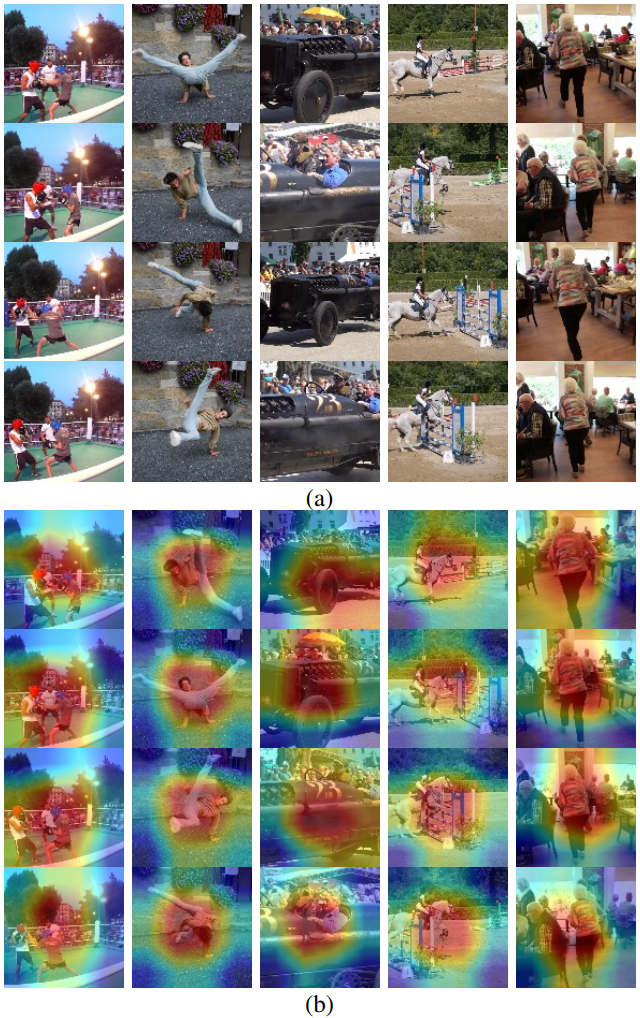}
	\caption{\textbf{Generalizability.} The figure shows the seven test sets (one column each) of unordered image sequences extracted from DAVIS dataset which has been correctly classified by the proposed approach when the network is trained on the dataset extracted from UCF-101. Each column shows one image set. (a) shows the unordered image sets which are given as the input to the proposed network. (b) shows the order of images provided by the proposed approach as the output along with the motion heat maps computed from the output of the last DAB of the network.}
	\label{fig:davis}
\end{figure}

\begin{figure*}
	\centering

	\includegraphics[width=\linewidth]{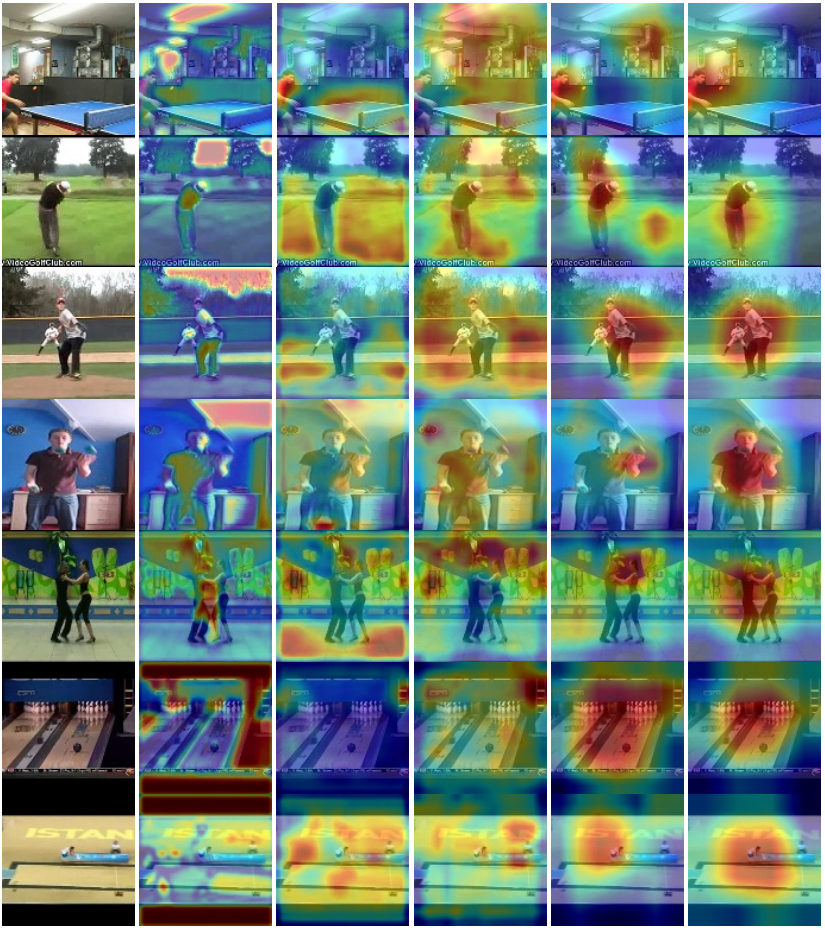}
	\caption{\textbf{Progression of motion heat maps.} The figure shows the progression of the motion heat maps computed using the output of DAB along the depth of the ResNet (50 layers) which is used as the backbone architecture for the proposed convolution block. The first column shows one of image of the input image sequences. The second column shows the output of DAB used after conv1 of ResNet \cite{he2016deep}. The third, fourth, fifth, and sixth columns show the outputs of the last DAB of layer1, layer2, layer3, and layer4 of the ResNet architecture, respectively}
	\label{fig:progression}
\end{figure*}
\subsubsection{Varying the number of images for accumulation of temporal differences }
In this study, we observe the effect of accumulating the differences of the feature map corresponding to each image with the feature maps of fixed number of images ahead of it in the input sequence. Let $\{\mathcal{F}_c^1,\mathcal{F}_c^2, \mathcal{F}_c^3,\ldots,\mathcal{F}_c^n\} \in \mathbb{R}^{c\times 1 \times h \times w}$ be the volumes along the temporal depth of $\mathcal{F}_c$. In this case, the output of DAB $\mathcal{F}_s \in \mathbb{R}^{c\times n \times h \times w} $ is obtained as shown in Eq. \ref{eq:DAB1}. Here, $\mathcal{F}_s$ is the concatenation of $\{\mathcal{F}_s^1,\mathcal{F}_s^2, \mathcal{F}_s^3,\ldots,\mathcal{F}_s^n \}\in \mathbb{R}^{c\times 1 \times h \times w}$ along the temporal depth.
\begin{equation}
\mathcal{F}_s^i = \sum\limits_{k=i}^{\min(i+m,n)}(\mathcal{F}_c^{i}-\mathcal{F}_c^{k}), \ \ \forall i = 2, \ldots, n-1
\label{eq:DAB1}
\end{equation}
Here, $i$ is a location along the temporal dimension and $m<n$. For $i=n$, $\mathcal{F}_s^n=\mathcal{F}_c^n$. Table \ref{tab:m_effect} shows the effect of varying the number of images used for accumulating the temporal differences, i.e., the value of $m$ in Eq. \ref{eq:DAB1}, by comparing the test classification accuracy (in percentage) obtained on the dataset of unordered image sequences of length 5 extracted from UCF-101 when trained with different values of $m$ (Eq. \ref{eq:DAB1}). For the sequence of length 5, there are $5!/2=60$ classes. It can be observed that without temporal differences, the network achieved the accuracy equivalent to the random probability of picking a class among 60 classes i.e. 0.0167. Even with the help of temporal differences computed among the adjacent images, the network still does not learn anything. It can be seen that as we increase the number of image used for the accumulation of temporal gradients, the test accuracy increases.
\subsubsection{Generalizability}
In this study, we evaluate the generalizability of the proposed approach. We want to verify that the proposed approach (ResNet-50 as the backbone) is learning the task of sequencing rather than somehow learning the distribution of the dataset. For this purpose, we use the dataset of image sequences extracted from the DAVIS dataset. We use the networks (ResNet50 as the backbone) trained on the dataset of sequence lengths 4 and 6 extracted from the UCF-101 to obtain the classification accuracy on the dataset of unordered image sequences of lengths 4 and 6 extracted from the DAVIS dataset.\\
Table~\ref{tab:davis} shows the classification accuracy obtained on the dataset of unordered image sequences extracted from DAVIS dataset. The dataset is extracted in such a way that the images in the sequence are evenly spaced temporally. However, the temporal spacing between the images could affect the classification accuracy of the temporal ordering. We extract different test sets of the image sequences from the DAVIS dataset by varying the number of frames skipped in the video while extracting the image sequences of length 4 and 6. Table~\ref{tab:davis} shows the variation of the classification accuracy as we change the temporal spacing between the images of the image sequence. It can be seen that as we increase the temporal spacing of the image sequences during the extraction, the classification accuracy decreases. This is because when the temporal spacing is large, the dynamic objects perform large motion which could lead to erroneous ordering. However, considering that we are evaluating the network on a dataset (in this case DAVIS dataset) which is different from the dataset is trained with (in this case UCF101), the obtained accuracy is considerably high. This shows that the proposed approach (ResNet50 as the backbone) learns the task of image sequencing. Fig. \ref{fig:davis} shows seven sets of unordered image sequences extracted from DAVIS dataset which have been correctly classified by the proposed network trained on the dataset extracted from UCF-101. Fig. \ref{fig:davis}(a) shows the unordered image sets which are given as the input to the proposed network. Fig. \ref{fig:davis}(b) shows the order of image set provided by the proposed network as the output.
\subsubsection{Motion Heat Maps}
We extract heat maps from the output of the last DAB of the network to understand the nature of the feature maps computed by DAB. DAB outputs the volumes of the accumulated temporal differences, i.e., $\{\mathcal{F}_s^1,\mathcal{F}_s^2, \mathcal{F}_s^3,\ldots,\mathcal{F}_s^n\}$ corresponding to each of the $n$ input images. We compute the heat map for the $i^{th}$ image by averaging the absolute values of the feature maps belonging to $\mathcal{F}_s^i$ along the channels. Fig. \ref{fig:ucf101} and \ref{fig:davis}(b) shows the motion heat maps obtained from the output of the last DAB of the network. It can be observed that the DAB is focusing in the regions with significant motion. It is significant as we have not provided any motion related cues to the network.
\subsubsection{Progression of motion heat maps}
In this study, we observe the progression of the motion heat maps computed using the output of DAB along the depth of the network. For the experiment, we used ResNet (50 layers) as the backbone architecture trained on the dataset of image sequences of length 6 extracted from UCF101. The architecture of ResNet is a sequence of conv1, layer1, layer2, layer3, layer4 and fc. Here, conv1 is the convolution layer and fc is the fully connected layer. layer1, layer2, layer3, and layer4 comprises of 3, 4, 6, and 3 residual blocks, respectively \cite{he2016deep}. We used the output of DAB used after conv1 and the outputs of the last DAB of layer1, layer2, layer3, and layer4 to demonstrate the progression of the motion heat maps. Figure \ref{fig:progression} shows the progression of the motion heat maps computed using the output of DAB along the depth of the ResNet (50 layers) which is used as the backbone architecture for the proposed convolution block. The first column shows one of the image of the input image sequences. The second column shows the output of DAB used after conv1 of ResNet \cite{he2016deep}. The third, fourth, fifth and sixth columns show the outputs of the last DAB of layer1, layer2, layer3, and layer4 of the ResNet architecture, respectively.

\section{Conclusion}
\label{sec:conclusion}
In this work, we propose a novel convolutional block for the task of image sequencing. We use residual network architecture as the back-bone for the proposed convolutional block \cite{he2016deep}. We outperform the state-of-the-art methods on the standard dataset used in the previous works by a significant margin. Through experiments, we verify the significance of the proposed difference accumulator block (DAB). We show through experiments that the sign of the differences of the feature maps holds an important information. We also show that the proposed approach generalizes well by evaluating it on DAVIS dataset on which the networks has not been trained. Generalizability has been a major concern in deep learning for a long time. The networks trained on one dataset does not perform well on the dataset which they have not been trained with even when the task is same and quite general like estimation of optical flow and semantic segmentation. The proposed approach is observed to overcome this issue for the task of image sequencing.


\bibliographystyle{IEEEtran}
\bibliography{references}
	
%


\end{document}